\documentclass[sigconf, anonymous=false]{acmart}

\AtBeginDocument{%
  \providecommand\BibTeX{{%
    \normalfont B\kern-0.5em{\scshape i\kern-0.25em b}\kern-0.8em\TeX}}}

\setcopyright{none}
\copyrightyear{2023}
\acmYear{2023}
\acmDOI{XXXXXXX.XXXXXXX}

\acmConference[Preprint]{}{currently under review}{}
\acmPrice{15.00}
\acmISBN{978-1-4503-XXXX-X/18/06}

\usepackage{multirow}
\usepackage{booktabs, bm}
\usepackage{subcaption}

\begin{document}

\title{FinDiff: Diffusion Models for Financial Tabular Data Generation}






\author{%
  \vspace{1em} 
  Timur Sattarov\textsuperscript{1,2}
  \hspace{6em}
  Marco Schreyer\textsuperscript{1}
  \hspace{5em}
  Damian Borth\textsuperscript{1} \\
}

\affiliation{%
  timur.sattarov@bundesbank.de
  \hspace{4em}
  marco.schreyer@unisg.ch
  \hspace{4em}
  damian.borth@unisg.ch  \hspace{1em} \\
  \vspace{0.5em} 
  \textsuperscript{1}University of St. Gallen, St. Gallen, Switzerland \hspace{2em}
  \textsuperscript{2}Deutsche Bundesbank, Frankfurt am Main, Germany \\
  \vspace{1em} 
  \country{}
}

\renewcommand{\shortauthors}{T. Sattarov, M. Schreyer, and D. Borth}

\begin{abstract}

The sharing of microdata, such as fund holdings and derivative instruments, by regulatory institutions presents a unique challenge due to strict data confidentiality and privacy regulations. These challenges often hinder the ability of both academics and practitioners to conduct collaborative research effectively. The emergence of generative models, particularly diffusion models, capable of synthesizing data mimicking the underlying distributions of real-world data presents a compelling solution. This work introduces 'FinDiff', a diffusion model designed to generate real-world financial tabular data for a variety of regulatory downstream tasks, for example economic scenario modeling, stress tests, and fraud detection. The model uses embedding encodings to model mixed modality financial data, comprising both categorical and numeric attributes. The performance of FinDiff in generating synthetic tabular financial data is evaluated against state-of-the-art baseline models using three real-world financial datasets (including two publicly available datasets and one proprietary dataset). Empirical results demonstrate that FinDiff excels in generating synthetic tabular financial data with high fidelity, privacy, and utility.

\end{abstract}




\maketitle

\begin{figure}[ht]
  \centering
  \includegraphics[width=\linewidth]{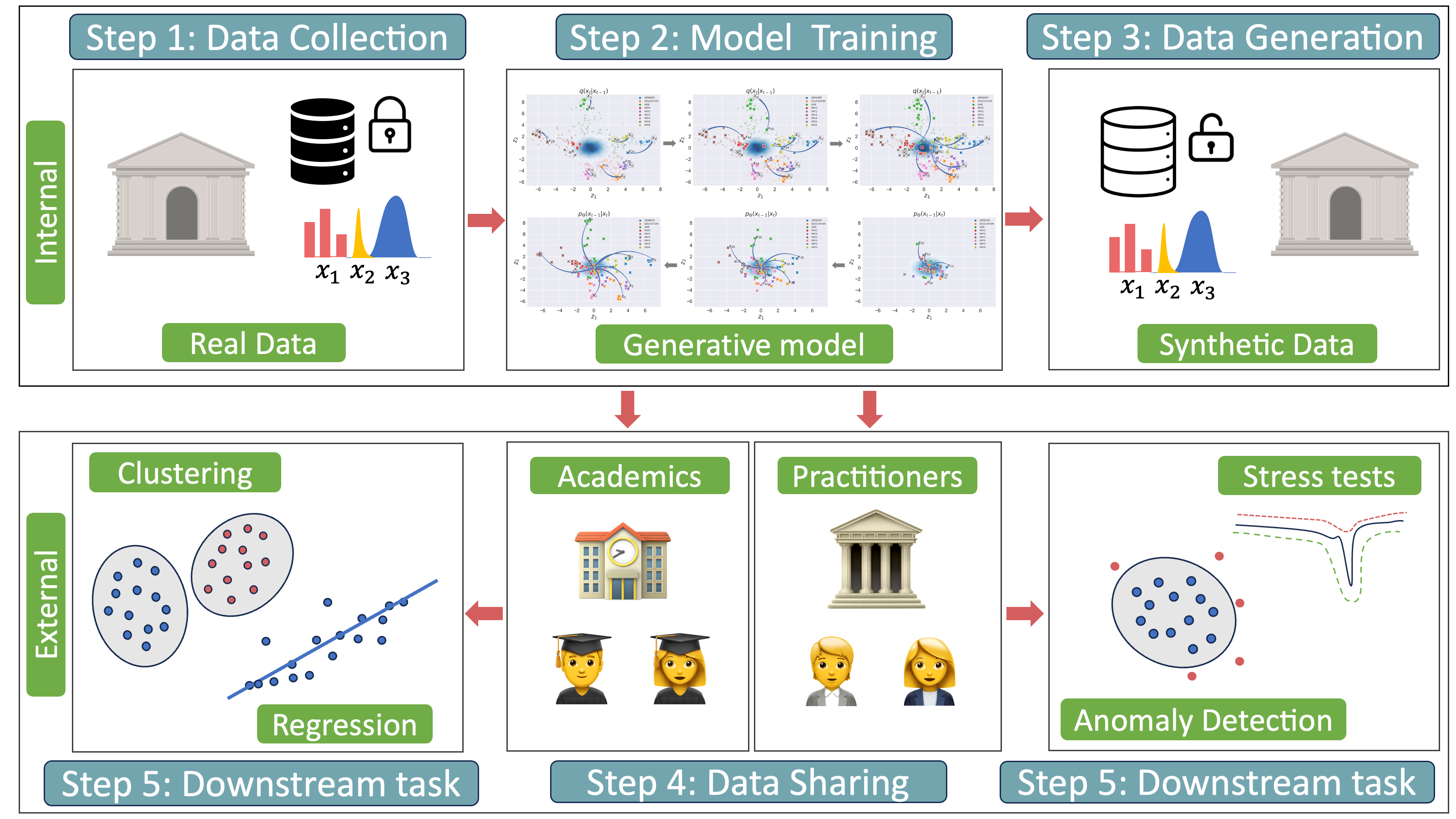}
  \caption{The process of synthetic data generation within the financial regulatory landscape and its subsequent dissemination amongst researchers and practitioners for advanced downstream applications.}
  \label{fig:data_sharing_process}
\end{figure}

\section{Introduction}

In the dynamically evolving financial regulatory landscape, data analytics plays an increasingly crucial role. Central banks worldwide, with a particular focus on Europe, amass substantial quantities of microdata. This data is pivotal for guiding policy decisions, conducting risk assessments, and ensuring financial stability. However, the granular nature of this data also engenders unique challenges.

At the forefront of these challenges are the stringent regulations surrounding data privacy, such as the General Data Protection Regulation (GDPR) in Europe. These regulations inhibit the dissemination of raw financial data, restricting not only independent researchers and practitioners who could employ this data for model training and knowledge discovery, but also hindering collaborative research efforts across different central banks.

In response to these challenges, several recent initiatives have underscored the importance of collaborative research. An example is the establishment of the Financial Big Data Cluster (FBDC)\footnote{\url{https://www.bmwk.de/Redaktion/EN/Artikel/Digital-World/GAIA-X-Use-Cases/financial-big-data-cluster-fbdc.html}}, a collaborative effort bringing together banks, regulatory institutions, Start-ups \& SMEs with Research and Development to build a cloud-based financial data pool for developing AI applications. Such clusters could serve as foundations for Financial Market Authorities in designing AI-based anti-money laundering systems.

Another important aspect is data sharing, which is the secure and efficient distribution of data resources across multiple users or organizations, using a mix of technologies and legal frameworks, designed to bolster collaboration and risk awareness. The Irving Fisher Committee's (IFC) recent guidance note\footnote{\url{https://www.bis.org/ifc/data_sharing_practices.pdf}} further emphasizes the significance of data sharing and collaboration among central banks. This note explores the challenges, benefits, and initiatives fostering cross-border cooperation, highlighting the potential for harmonizing data practices and leveraging advanced technologies within the financial sector. Furthermore, the United Nations Economic Commission for Europe (UNECE) has recently published the guide on data sharing in official statistics\footnote{\url{https://unece.org/sites/default/files/2021-02/Data\%20sharing\%20guide\%20on\%20web_1.pdf}}. One of the key aspects to optimize economic statistics was to broaden the data scope for national statistical offices by leveraging statistics from international authorities, moving beyond their own national data. 

In this context, synthetic data generation offers a compelling resolution to these challenges. Synthetic data, mirroring the statistical properties of original data without compromising sensitive information, can broaden the accessibility of essential data for research and machine learning purposes. It facilitates financial model testing, enhances transparency, nurtures collaborative research across institutions, and ensures data privacy regulation compliance. Moreover, synthetic data can simulate diverse economic scenarios, assisting in stress tests and other predictive tasks vital to regulatory operations. 
One area within AI that holds great promise for finance is generative AI, which encompasses the learning of models to generate new synthetic data with high fidelity\citep{synth_data_finance}.

Building upon this motivation, our work introduces a diffusion model designed for synthesizing financial tabular data. Originally developed for image synthesis tasks \citep{diffusion_models_survey, zhang2023texttoimage}, diffusion models have shown remarkable results in various fields, including natural language processing \citep{zou2023survey} and audio \citep{chen2020wavegrad, kong2021diffwave}. Diffusion models simulate a random walk to gradually transform data from a simple initial distribution to the complex distribution of the original data. Due to their unique capabilities to model high-dimensional dependencies, they are well-positioned to address the challenge of data sharing in a regulatory environment. By employing these models, banks can generate synthetic data that maintains the complex statistical properties of the original data, without disclosing any sensitive information. This allows the models to maintain the usefulness of the data for machine learning applications while also ensuring compliance with data privacy regulations.

In summation, this study makes the following contributions:

\begin{itemize}
\item The development and implementation of a diffusion model, named FinDif, which capably synthesizes real-world financial tabular data. 
\item The utilization of embedding encoding for addressing the challenges inherent to mixed modality financial data, thus fostering efficient data representation. 
\item The rigorous assessment of the quality of the data generated by the model using fidelity, privacy, and utility metrics, attesting to the model's precision and effectiveness.
\end{itemize}

The remainder of this paper is structured as follows: \autoref{sec:related_work} provides an overview of the related work. In \autoref{sec:methodology} we describe
the diffusion model basics and outline the proposed methodology, FinDiff. Next, \autoref{sec:experimental_setup} and \autoref{sec:experimental_results} outline the experimental setup and results. We conclude the paper with a summary and future
research directions in \autoref{sec:conclusion}.

\section{Related Work}
\label{sec:related_work}

The literature survey hereafter focuses on (1) existing Gaussian diffusion models, and (2) developed models designed for the generation of financial data.

\subsection{Gaussian Diffusion Models}

The initial development of diffusion models can be attributed to Sohl-Dickstein et al.\citep{sohldickstein2015deep} which was followed by a number of improvements and variations like Denoising Diffusion Probabilistic Models (DDPM)\citep{ho2020denoising}, Denoising Diffusion Implicit Models (DDIM) with accelerated sampling)\citep{song2022denoising}. Another improvement was introduced by Nichol et al. \citep{nichol2021improved} where they compared the coverage of real data by DDPMs and GANs. Rombach et al. showed that with Latent Diffusion Models (LDM)\citep{rombach2022highresolution} one
can generate high-quality image synthesis while significantly reducing computation time.

\subsection{Discrete Diffusion Models}

Strudel et al. \citep{strudel2022selfconditioned} introduced Self-conditioned Embedding Diffusion (SED), where they successfully showed the applicability of an embedding-based diffusion model for text generation. They also showed that SED can rival Autoregressive models.
Recently, Gao et al. \citep{gao2023difformer} applied embeddings on the discrete text data generation and showed a number of training challenges such as collapse of the denoising objective or imbalanced scales of the embedding norms. 
Another example of diffusion models for text generation is the Diffusion-LM introduced by Li et al. \citep{li2022diffusionlm}, however here the authors focused on controlling complex, fine-grained outputs.

\begin{figure*}[ht!]
  \centering
  \includegraphics[width=\linewidth, keepaspectratio]{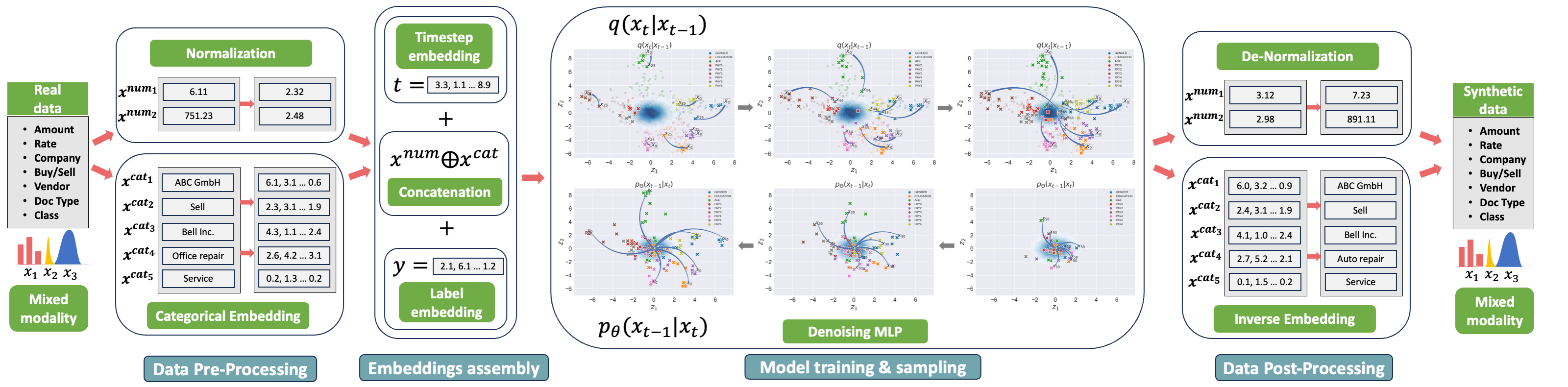}
  \caption{Schematic diagram of the 'FinDiff' model. The process begins with the pre-processing phase, where categorical attributes are transformed into embeddings and numerical attributes undergo normalization. During the embedding assembly phase, these processed data streams combine into an aggregate $x_0=e^{c_1} \oplus e^{c_2} \oplus ... \oplus e^{c_k} \oplus x^{num}$, further enriched with time and label embeddings. This combined data is then passed into a Feed Forward Neural Network to estimate the added noise component. Once the model is trained and new data is generated via Gaussian sampling, the post-processing phase maps the resultant embedding $\hat{x^{cat_i}}$ back to its nearest counterpart. Concurrently, numerical attributes are denormalized, restoring to their original data space.}
  \label{FinDif_sketch}
\end{figure*}

\subsection{Financial Data Generation}

The generation of synthetic financial data has been a topic of interest in recent years. Wiese et al. \citep{wiese2019quant} introduced Quant GANs, a data-driven model that utilizes temporal convolutional networks (TCNs) to capture long-range dependencies such as the presence of volatility clusters. Ni et al. \citep{ni2020conditional, ni2021sigwasserstein} developed high-fidelity time-series generators, the SigWGAN, by combining continuous-time stochastic models with the newly proposed signature W1 metric. Their proposed model was validated on both synthetic data generated by popular quantitative risk models and empirical financial data.
Dogariu et al. \citep{dogariu2021synthetic} proposed several solutions for augmenting financial datasets by synthesizing realistic time-series using generative models. 

Tabular data is another popular data modality where synthetic data generation is gaining momentum \citep{fonseca_tabular_2023}. Here, Variational Autoencoders \citep{vae2014} are perhaps one of the most popular models used for tabular data generation \citep{vae_fin1, vae_tab1, vardhan2020generating}. Xu et al.\citep{tvae_ctgan} have proposed TVAE specifically designed for tabular data generation tasks. In the same paper, they introduced CTGAN, Generative Adversarial Based model for tabular datasets. Another set of GAN-based models \citep{engelmann2021conditional, fan2020relational, schreyer2019adversarial} is also an active research direction in this community. The recent attempt to model tabular data using diffusion models was done by Kotelnikov et al. \citep{kotelnikov2022tabddpm}. However, since the categorical attributes were modeled using multinomial diffusion models \citep{argmax_flows}, they had to use one-hot encoding transformation, which has its disadvantage for large datasets. Another recent attempt to model tabular data with diffusion models is \textit{MissDiff} \citep{ouyang2023missdiff} which is capable of training synthesizer directly on the data with missing values.

 \vspace{0.5em}

To the best of our knowledge, this is the first attempt to develop a diffusion model for synthesizing financial tabular data using embedding encoding for categorical attributes.

\section{Methodology}
\label{sec:methodology}

In this section, we describe the Gaussian Diffusion Models and proposed Financial Diffusion (FinDiff).

\subsection{Gaussian Diffusion Models}
Denoising diffusion probabilistic model \citep{sohldickstein2015deep}, \citep{ho2020denoising} is a latent variable model that utilizes a forward process to perturb the data $x_0 \in \mathbb{R}^d$  step by step with Gaussian noise $\epsilon$, and then restore the data back in the reverse process. The forward process is started at $x_0$ and latent variables $x_1 ... x_T$ are generated with a Markov Chain by gradually perturbing it into a pure Gaussian noise $x_T \sim \mathcal{N}(\textbf{0,I})$. Hence, each Markov transition has the form: 

\begin{equation}
    q(x_t|x_{t-1})=\mathcal{N}(x_t;\sqrt{1-\beta_t}x_{x-1}, \beta_t, \textbf{I})
    \label{eq:q_step}
\end{equation}

\noindent where $\beta_t$ is the noise level added at timestep $t$. Sampling $x_t$ from $x_0$ for an arbitrary $t$ can also be achieved in a closed form $q(x_t|x_0)=\mathcal{N}(x_t;\sqrt{1-\hat{\beta_t}}x_0, \hat{\beta_t}, \textbf{I})$ where $\hat{\beta_t}=1-\prod_{i=0}^{t} (1-\beta_i)$. 

In the reverse process, the model incrementally denoises the latent variables $x_t$ to recover the data $x_0$. To approximate this process, we train a neural network with parameters $\theta$ and each denoising step is parameterized as:

\begin{equation}
    p_\theta(x_{t-1}|x_t)=\mathcal{N}(x_{t-1}; \mu_\theta(x_t, t), \Sigma_\theta(x_t,t))
    \label{eq:p_step}
\end{equation}

\noindent where $\mu_\theta$ and $\Sigma_\theta$ are the estimated mean and covariance of $q(x_t|x_{t-1})$. Since $\Sigma_\theta$ is diagonal, as suggested by Ho et al.\citep{ho2020denoising}, the computation of $\mu_\theta$ is the following:

\begin{equation}
    \mu_\theta(x_t, t)=\frac{1}{\sqrt{\alpha_t}}(x_t - \frac{\beta_t}{\sqrt{1-\hat{\alpha_t}}} \epsilon_\theta(x_t, t))
\end{equation}

\noindent where $\alpha_t:=1-\beta_t$, $\hat{\alpha_t}:=\prod_{i=0}^{t}\alpha_i$ and $\epsilon_\theta(x_t, t)$ is the predicted noise component. It was empirically shown that simplified loss of mean-squared errors between the ground truth $\epsilon$ and estimated $\epsilon(x_t, t)$ empirically leads to better results than tractable variational lower bound $log\; p_{\theta}(x_0)$:

\begin{equation}
    \mathcal{L}_t=\mathbb{E}_{x_0,\epsilon,t}||\epsilon-\epsilon_\theta(x_t,t)||_2^2
    \label{eq:loss}
\end{equation}

Though such a framework works well on continuous data, such as images, it cannot be directly applied to discrete data, such as categorical attributes of tabular data.

\subsection{Financial Diffusion (FinDif)}

\begin{figure*}[ht]
    \centering
    \begin{subfigure}{\textwidth}
    \includegraphics[width=\linewidth]{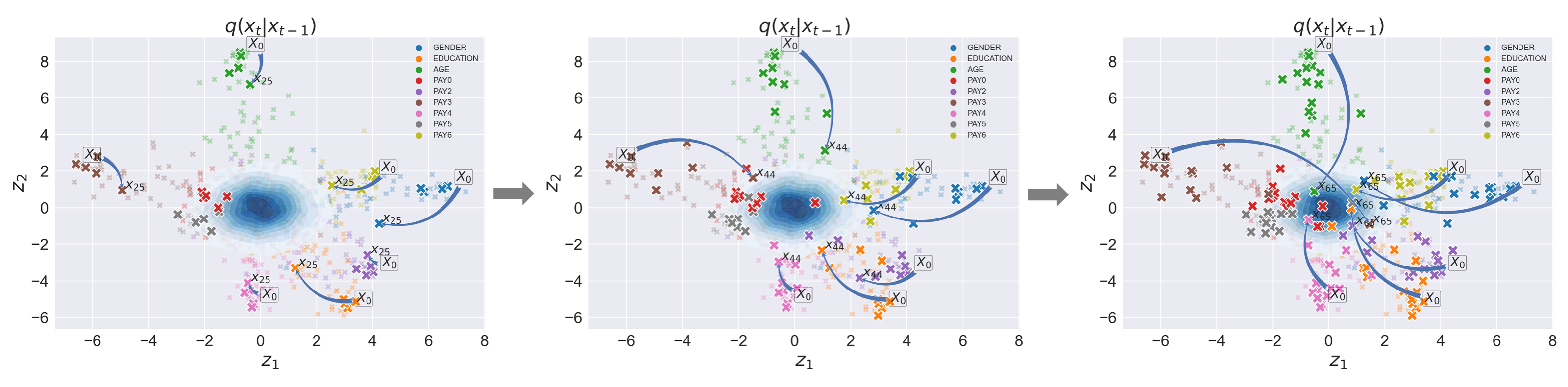}
    \caption{The forward diffusion process of three randomly selected noisy batches. This illustration presents the navigational path towards the Gaussian center, produced by gradually perturbing $x_0$ with Gaussian noise, culminating in the final noisy data representation $x_T \sim \mathcal{N}(0,I)$. In this step, each embedding representation converges towards the center of Gaussian $x_T \sim \mathcal{N}(0,I)$.}
    \label{fig:forward_process}
    \end{subfigure}

    \begin{subfigure}{\textwidth}
    \centering
    \includegraphics[width=\linewidth]{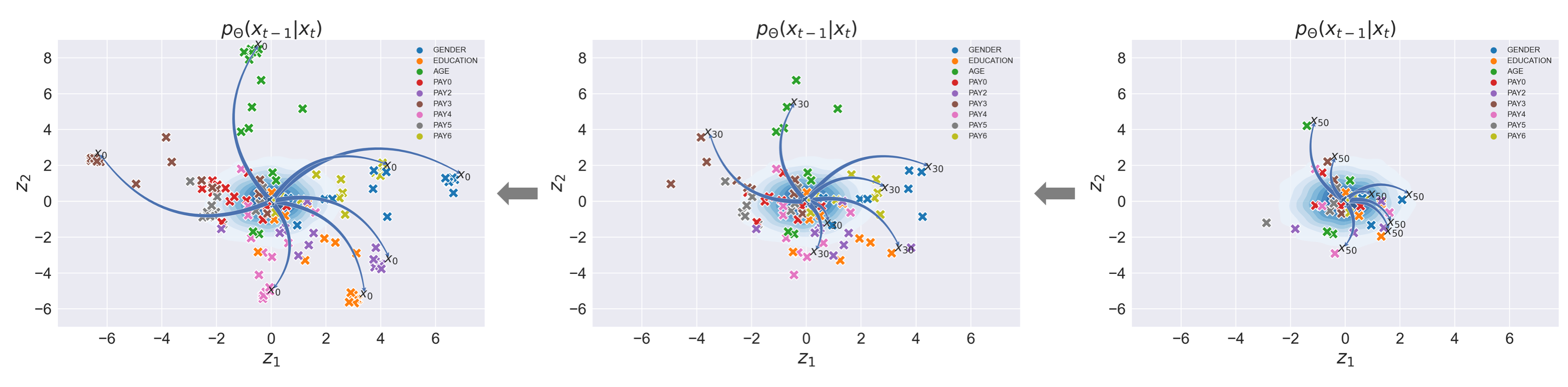}
    \caption{Three snapshots of the reverse diffusion process. The model employs this path to drift back towards the initial embedding representation $x_0$. In this step, the embedding drifts back in the direction of the starting token $X_0$}
    \label{fig:backward_process}
    \end{subfigure}

\caption{Example of the diffusion forward and reverse processes of 9 categorical embeddings of a single observation from the Credit Default dataset. Each cross point symbolizes the 2-dimensional embedding of a specific category, either at its origin $x_0$ or subsequent noisy representations such as $x_{25}$, $x_{44}$, $x_{65}$. The blurry points symbolize the noisy representations that remain unseen by the model.}
\label{fig:diff_process}

\end{figure*}

Our approach addresses the challenges inherent to mixed modality financial tabular data, comprising both categorical and numeric attributes. The categorical data, discrete variables exhibiting a finite set of possible values, has often been a stumbling block for conventional numerical models. We transform this categorical data, denoted $x^{cat}=(x^{c_1},x^{c_2},...,x^{c_N})$, into semantically rich, continuous representations via the use of embeddings, $e\in \mathbb{R}^D$. Each categorical element, $x^{c_i}$, is thus represented in a condensed vector space, positioning similar categories in close proximity, yielding an embedding matrix $E\in \mathbb{R}^{D\times C}$.

As illustrated in \autoref{FinDif_sketch}, the proposed methodology encompasses several stages. In the data pre-processing phase, numeric attributes undergo normalization while categorical tokens are embedded using a matrix $E$, which then, in the embedding assembly phase, are concatenated and aggregated with time and label embeddings. Gaussian noise is subsequently introduced according to \autoref{eq:q_step}, marking a diffusion step $\mathbf{t}$. To compute the time embedding $\mathbf{t}$, a sinusoidal position embedding is utilized, in alignment with Nichol et al.'s methodology \citep{nichol2021improved}. Both time embeddings and encoded sample $x_0$ undergo a linear layer before being aggregated, following the approach of Nichol et al. \citep{nichol2021improved}. The representation thus obtained is channeled through a Feed Forward Neural Network, where the model strives to estimate the added noise component, as delineated in \autoref{eq:loss}.
Once the model is trained, the sampled new data in the post-processing phase is mapped back to its respective category using the closest representation within the embedding matrix $E$, while numerical attributes are denormalized.

In \autoref{fig:diff_process}, the model's training and sampling steps are further described, utilizing a real-world example. This figure depicts the learning trajectory of categorical embeddings from a single observation, alongside the subsequent sampling step. In \autoref{fig:forward_process}, the forward process is illustrated, showing the navigation path towards the Gaussian center $\mathcal{N}(0,I)$. During the reverse process, demonstrated in \autoref{fig:backward_process}, the model employs this path to drift back towards the initial embedding representation $x_0$.  This arrangement enables the model to perceive the latent structure and interrelations within the categorical features, thereby fostering meaningful correlations during sampling.

\section{Experimental Setup}
\label{sec:experimental_setup}

In this section, we describe the details of the conducted experiments.
We describe the datasets as well as the data preprocessing steps, together with diffusion model training setup, baselines and the evaluation metrics.
For training and evaluation of the models, the PyTorch v1.13.0 \cite{pytorch} framework was used. For evaluation metrics as well as the implementation of baseline algorithms TVAE and CTGAN, Synthetic Data Vault (SDV) library v1.0.1 \citep{SDV} was used.

\subsection{Datasets and Data Preparation}

We benchmark the developed technique with three real-world datasets. 
Below, we provide the description of each dataset: 

\begin{itemize}
    \item \textbf{Credit Default}\footnote{The dataset is publicly available via: \url{https://archive.ics.uci.edu/ml/datasets/default+of+credit+card+clients}}: The dataset comprises bill statements of credit card clients, their default payments, history of payment as well as the demographic factors of the clients in Taiwan during the period April 2005 to September 2005 \citep{credit_default}. 

    \item \textbf{City of Philadelphia}\footnote{The dataset is publicly available via:\url{https://data.phila.gov/visualizations/payments}}: The dataset contains checks and direct deposit payments made by the City of Philadelphia during 2017 fiscal year. 
    
    \item \textbf{Fund Holdings}\footnote{In compliance with strict data privacy regulations, neither content nor the descriptive statistics of the dataset can be made publicly available.}: This proprietary dataset consists of the individual holdings of the investment funds issued by investment companies \citep{Holdings_hidden}. Each record reflects the asset or liability value submitted by the reporting entity at the end of the month. 
\end{itemize}

\begin{table}[h]
\caption{\label{tab:datasets} Descriptive statistics of the selected datasets}
\begin{tabular}{lrrrr}
\toprule

\multirow{2}{*}{Data}  &  \multirow{2}{*}{Rows} & \multicolumn{2}{c}{Columns} & \multirow{2}{*}{Classes} \\
                        &                     & Categ. & Num.  &  \\

\midrule
Credit Default  & 30,000    & 10   & 13 & 2 \\
Philadelphia Payments & 100,000 &   7   &   1   &   11  \\
Fund Holdings       & 88,893   & 6   & 78 & 18 \\
\bottomrule
\end{tabular}
\end{table}

In considering dataset selection, we prioritized diversity in the proportion of categorical and numeric attributes. Specifically, the Philadelphia Payments dataset is primarily composed of categorical attributes, the Fund Holdings dataset predominantly contains numeric attributes, and the Credit Default dataset presents a nearly equal distribution of both attribute types, as \autoref{tab:datasets} attests. The numeric attributes were normalized to the zero mean and unit variance. Notably, we were constrained to sample only 100,000 observations from the total 238,894 in the Philadelphia dataset due to limitations faced during the training of baseline models. The need for a one-hot encoded representation for these models led to exceedingly high-dimensional transformed datasets of 6,124 (7,830) dimensions. Unlike these baseline models, our proposed method, FinDif, utilizes embeddings instead of one-hot encoding, thereby circumventing such dimensionality issues.

\subsection{Diffusion Model Training Setup}

Every dataset was split into training and test sets by a fraction of 70 and 30 correspondingly. All models are trained on the train set and the evaluation metrics are collected with respect to the test set.

\noindent \textbf{Architecture Setup.} The precise network architecture utilized for each dataset was selected after the exploration of the hyperparameter space. We determined the most appropriate number of diffusion steps to be 500, using a linear scheduler. Each categorical attribute was assigned a dimensionality of 2, which proved beneficial for both visual inspection (see \autoref{fig:diff_process}) and computational efficiency. Moreover, our empirical observations indicated that increasing the dimensionality of the categorical embeddings did not result in a performance enhancement. Strudel et al. \citep{strudel2022selfconditioned} similarly reported that a high-dimensional embedding space leads to performance degradation when training such models on text data. 

\noindent \textbf{Hyperparameters.} Depending on the dataset, a distinct combination of the number of neurons and layers was selected. The most effective hyperparameters were identified following an exhaustive search across the following space: number of neurons [256, 512, 1024, 2048, 4096, 8192] and number of hidden layers [4, 6, 8, 10, 12]. We train every model for a maximum of 3000 epochs with a mini-batch of size 512 and use the Adam optimizer \citep{adam} with $\beta_{1}=0.9$, $\beta_{2}=0.999$ in combination with a cosine learning rate scheduler. The learning weights are randomly initiated as described in \citep{Glorot10understandingthe}.

\noindent \textbf{Baselines.} For benchmarking the performance of the FinDiff we compare it against three baseline models.

\begin{itemize}
    \item \textbf{TVAE}  \citep{tvae_ctgan} - adapted version of the variational autoencoder, specifically designed for mixed-type tabular data generation.
    \item \textbf{CTGAN} \citep{tvae_ctgan} - GAN-based synthetic tabular data generator, shown to perform well on tabular datasets. 
    \item \textbf{TabDDPM} \citep{kotelnikov2022tabddpm} - recently developed diffusion-based alternative for generation of tabular datasets.
\end{itemize}

\subsection{Evaluation Metrics}

\begin{table*}[ht]
  \centering
    \caption{Comparative analysis of different synthetic data generation models on various datasets. Each model is evaluated on several key parameters including Fidelity (Column and Row), Utility, Synthesis, and Privacy. Scores presented are averages with standard deviations from 5 experiments of random seeds. Boldface numbers indicate the best-performing model for a given measure on a particular dataset.}
      \label{tab:quant_results}
  \begin{tabular}{@{}l l c c c c c@{}}
    \toprule
    & & \multicolumn{5}{c}{\textbf{Evaluation Measures}} \\
    \cmidrule{3-7}
    \textbf{Dataset} & \textbf{Model} & \textbf{Fidelity Column $\uparrow$} & \textbf{Fidelity Row $\uparrow$} & \textbf{Utility $\uparrow$} & \textbf{Synthesis $\uparrow$} & \textbf{Privacy $\downarrow$}\\
    \midrule
    \multirow{4}{*}{Credit Default} & TVAE & 0.920 $\pm$ 0.01 & 0.923 $\pm$ 0.01 & 0.790 $\pm$ 0.01 & 1.000 $\pm$ 0.00 & 1.573 $\pm$ 0.01\\
     & CTGAN & 0.872 $\pm$ 0.01 & 0.871 $\pm$ 0.01 & 0.703 $\pm$ 0.03 & 1.000 $\pm$ 0.00 & 1.880 $\pm$ 0.06\\
     & TabDDPM & 0.401 $\pm$ 0.05 & 0.320 $\pm$ 0.01 & 0.709 $\pm$ 0.02 & 1.000 $\pm$ 0.00 & 2.734 $\pm$ 0.18\\
     & FinDiff & \textbf{0.931} $\pm$ \textbf{0.01} & \textbf{0.939} $\pm$ \textbf{0.01} & \textbf{0.794} $\pm$ \textbf{0.01} & 1.000 $\pm$ 0.00 & \textbf{1.474} $\pm$ \textbf{0.01}\\

    \midrule
     \multirow{4}{*}{Philadelphia Payments} & TVAE & 0.791 $\pm$ 0.01 & 0.634 $\pm$ 0.01 & 0.785 $\pm$ 0.03 & 0.992 $\pm$ 0.01 & 2.000 $\pm$ 0.00\\
     & CTGAN & 0.782 $\pm$ 0.01 & 0.576 $\pm$ 0.01 & 0.728 $\pm$ 0.01 & 0.994 $\pm$ 0.01 & 1.765 $\pm$ 0.32\\
     & TabDDPM & 0.900 $\pm$ 0.01 & 0.535 $\pm$ 0.01 & 0.863 $\pm$ 0.01 & \textbf{1.000} $\pm$ \textbf{0.01} & 4.135 $\pm$ 0.20\\
     & FinDiff & \textbf{0.901} $\pm$ \textbf{0.01} & \textbf{0.838} $\pm$ \textbf{0.00} & \textbf{0.874} $\pm$ \textbf{0.00} & 0.992 $\pm$ 0.01 & \textbf{1.414} $\pm$ \textbf{0.00}\\     

    \midrule
     \multirow{4}{*}{Fund Holdings} & TVAE & 0.745 $\pm$ 0.01 & \textbf{0.952} $\pm$ \textbf{0.01} & 0.543 $\pm$ 0.02 & 1.000 $\pm$ 0.00 & \textbf{0.171} $\pm$ \textbf{0.01}\\
     & CTGAN & 0.591 $\pm$ 0.02 & 0.937 $\pm$ 0.01 & 0.421 $\pm$ 0.03 & 1.000 $\pm$ 0.00 & 0.847 $\pm$ 0.29\\
     & TabDDPM & 0.119 $\pm$ 0.01 & 0.767 $\pm$ 0.01 & 0.135 $\pm$ 0.06 & 1.000 $\pm$ 0.00 & 9.816 $\pm$ 8.00\\
     & FinDiff & \textbf{0.764} $\pm$ \textbf{0.01} & 0.949 $\pm$ 0.01 & \textbf{0.544} $\pm$ \textbf{0.02} & 1.000 $\pm$ 0.00 & 3.667 $\pm$ 0.40\\ 
    \bottomrule
  \end{tabular}
\end{table*}

To assess the quality of generated data, we evaluate the excellence from various viewpoints: fidelity, privacy, utility, and synthesis.

\noindent \textbf{Fidelity.} To quantify the relevance of generated data, we evaluate it in terms of fidelity. Fidelity assesses how well the synthetic data mimics the real data. This is performed on the column as well as on the row levels.  The column fidelity is examined using the similarity of every column in the synthetic dataset against the real dataset. To measure it for numeric attributes we use Kolmogorov-Smirnov statistic $KS(x^d, s^d)$ \citep{kolmogorov_1951} which reflects the maximum difference between an empirical and hypothetical cumulative distribution. For categorical attributes, we utilize the Total Variation Distance between the real and synthetic columns, defined as $\textit{TVD} (x^d, s^d) = \sum_{c\in C} | p(x^{d_c})- p(s^{d_c})|$ where $p(s^{d_c})$ is the fraction of existing categories $\mathbf{c}$ in the attribute $d$. Therefore, the column-wise fidelity ($\omega_{col}$) obtains the following form: 

\begin{equation}
    \omega_{col}(x^d, s^d)=\begin{cases}
        1-KS(x^d, s^d) & \text{if $d$ is numerical} \\
        1-\frac{1}{2}TVD(x^d, s^d) & \text{if $d$ is categorical} 
    \end{cases}
\end{equation}

\noindent The total score for the synthetic dataset $S$ is computed as the average across all attributes: $\Omega_{col}(X,S)=\frac{1}{D} \sum_{d=0}^D \omega(x^d, s^d)$. 

\noindent The row fidelity is measured using correlations between a pair of columns. For numeric attributes, a Pearson correlation \citep{benesty2009pearson} is computed between a pair of columns $\rho(x^a, x^b)=\frac{cov(x^a, x^b)}{\sigma(x^a)\sigma(x^b)}$. For categorical attributes, we used the contingency table by computing the Total Variation Distance on a pair of categories in attributes $A$ and $B$.  

\begin{equation}
    \omega_{row}(x^{a,b}, s^{a,b})=\begin{cases}
        1-\frac{1}{2}|\rho(x^a, x^b) - \rho(s^a, s^b)| & \text{if $a$,$b$ are numerical} \\
        1-\frac{1}{2} TVD(x^{a,b}, s^{a,b}) & \text{if $a$,$b$ are categorical} 
    \end{cases}
\end{equation}

\noindent The final fidelity score for a synthetic dataset $S$ is computed as the average score across all attribute pairs: 
\newline $\Omega_{row}(X,S)=\frac{1}{D} \sum_{d=0}^D \omega(x^{a,b}, s^{a,b})$.

\noindent \textbf{Privacy}. The privacy metric represents the extent to which the synthetic data prevents identification of the original data entries. To measure the privacy of the generated, we use the Distance to Closest Records (DCR). The DCR is computed as the closest distance from the synthetic data point $s_n$ to the real data points $X$, formally defined as:

\begin{equation}
    DCR(s_n) = \min_{x\in X} d(s_n, x)
\end{equation}

\noindent where $d(\cdot)$ is the distance metric (Euclidean is our case). The final score is computed as the median of DCRs of all synthetic data points.

\noindent \textbf{Utility Assessment.} A critical aspect of synthetic data evaluation is the measurement of utility, essentially quantifying its functional equivalence to the original data. Often referred to as Machine Learning efficacy, this evaluation involves training a machine learning model on the synthetic dataset, followed by its evaluation on the original dataset. This measurement criterion serves to assess the applicability and quality of the generated data for downstream tasks, such as classification or anomaly detection. We quantify utility by training a classifier on synthetic data of identical dimensions as the training set and subsequently evaluating it on the real test set. The utility is then represented by the mean accuracy, aggregated across all classifiers $\Phi = \frac{1}{5} \sum_{i=1}^{5} \Theta_i(S_{train}, X_{test})$. This study incorporates a diverse selection of 5 classifiers, including Random Forest, Decision Trees, Logistic Regression, Ada Boost, and Naive Bayes.

\noindent \textbf{Synthesis} Synthesis reflects the ability of a model to generate synthetic data that is not an exact replication of the real data. It checks whether the generated records are novel or exactly match the records in the original dataset. It is computed as the fraction of the matched synthetic records to the total number of generated records. A numeric entry is considered as a match if its value is within $1\%$ range of the real value.

\section{Experimental Results}
\label{sec:experimental_results}

In this section, we describe the results of the conducted experiments. We demonstrate the efficiency of the proposed technique, providing quantitative results with the ablation study and qualitative assessment.

\subsection{Quantitative Results}

\begin{figure*}[ht]
  \centering
  \includegraphics[width=\linewidth]{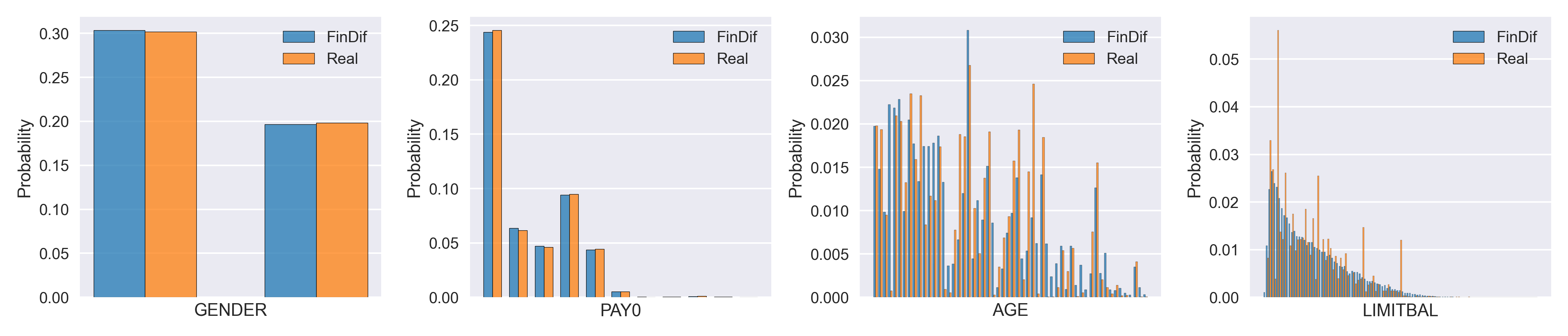}
  \caption{Feature distributions of the real and synthetic data generated by FinDiff model. The orange bars mimic the real-world data distribution, and the blue bars are the synthetic data generated by FinDiff model. Categorical features are "PAY0", "AGE" and "GENDER",  the numeric feature is "LIMITBAL".}
  \label{fig:distributions}
\end{figure*}

To quantitatively evaluate the developed technique, we assess the model from different perspectives. \autoref{tab:quant_results} shows the results of all metrics across three datasets and all baseline models.

\noindent \textbf{Fidelity.} For the Credit Default dataset, the FinDiff model outperformed all other models in terms of both columnar and row fidelity and achieved an approximately 1.2\% and 1.6\% increase in performance for these metrics compared to the second-best performing model, TVAE. The high fidelity scores suggest that FinDiff accurately captures both individual and joint distributions, which is crucial in preserving the integrity of the dataset's statistical properties. On the Philadelphia Payments dataset, the FinDiff model again demonstrated superior performance. On the Fund Holdings dataset, the FinDiff outperformed all models on column-wise metric and is the second-based model on the row-wise fidelity with a very small margin. We believe, that the reason lies in the nature of the Fund Holding dataset, where most of the numeric attributes are highly skewed with extremely high values. 

\noindent \textbf{Privacy.} In the context of the Credit Default dataset, the FinDiff model establishes a strong performance by yielding the lowest privacy score, measured at 1.474. This outstrips the performances of TVAE, CTGAN, and TabDDPM models by 6.28\%, 21.49\%, and 46.09\% respectively. This finding indicates that FinDiff can generate synthetic credit default data while ensuring superior privacy preservation. As for the Philadelphia Payments dataset, FinDiff continues its lead, returning the lowest privacy score. Analyzing the Fund Holdings dataset, we encounter an anomalous trend, where the TVAE model records the best privacy score at 0.171. In conclusion, the FinDiff model demonstrates robust privacy performance in the Credit Default and Philadelphia Payments datasets, significantly outperforming other models. While it cedes to the TVAE model in the Fund Holdings dataset, it still offers enhanced privacy protection relative to the TabDDPM model. These results underline that the FinDiff model generally excels at ensuring privacy preservation when generating synthetic data, although the efficiency of the privacy protection can be influenced by dataset-specific factors.

\noindent \textbf{Utility.} Starting with the Credit Default dataset, the FinDiff model demonstrates superior utility, with a score of 0.794. This surpasses the performances of TVAE, CTGAN, and TabDDPM models by 0.51\%, 12.95\%, and 11.98\% respectively, indicating that FinDiff optimally retains the essential characteristics of the original data. Moreover, when computing the utility accuracy using only real data (trained and tested on the real data), it yields only 0.706 which is 8.4\% lower than using the synthetic data of FinDif. This significantly confirms the usefulness and importance of strong synthesizers. For the Philadelphia Payments and Fund Holdings datasets, FinDiff continues to demonstrate its leading performance, although exceeding the second-based model by a small margin.

\noindent \textbf{Synthesis.} Analyzing synthesis scores across the three datasets, the models display generally strong performances. In the Credit Default and Fund Holdings datasets, all models, including FinDif, achieved perfect scores of 1.000. For the Philadelphia Payments dataset, while the TabDDPM model secured a perfect score, FinDif's performance was only marginally lower at 0.992. 

 \vspace{0.5em}

Upon examining the data from all three datasets, it is clear that FinDiff consistently delivered high scores across all evaluation measures. It showed exceptional performance in terms of Fidelity (both column and row), with peak scores observed in the Credit Default dataset. Furthermore, FinDiff demonstrated admirable privacy and utility numbers, evidenced by the lowest privacy scores for both the Credit Default and Philadelphia Payments datasets.  While the model's performance is not the leading one in the Fund Holdings dataset, it still maintains a significant advantage over CTGAN and TabDDPM models. This underscores the generally strong performance of the FinDiff model, though the degree of this performance can be contingent upon dataset-specific characteristics.

\subsection{Qualitative Results}

In order to gauge the quality of the data generated, feature distributions of the original and synthetic data of the FinDiff model are plotted, which reflect the column-wise fidelity capabilities. Figure \ref{fig:distributions} illustrates examples of three categorical features and a numerical one. In an effort to more effectively assess the quality of sampling, categorical attributes with varying quantities of unique categories are selected, such as 'GENDER'=2, 'PAY0'=11, and 'AGE'=56. The capacity of the FinDiff model to faithfully replicate the distributions of the original data is confirmed through this process.

\begin{figure*}[h]
  \centering
  \includegraphics[width=\linewidth]{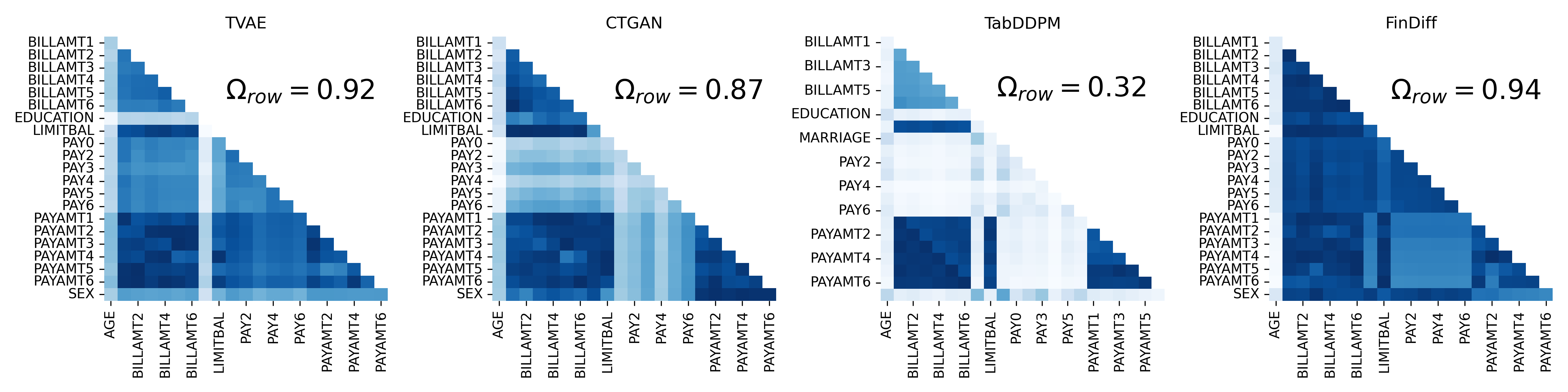}
  \caption{Absolute difference of feature correlations between the synthetic and original data generated by all models. A more intensive color gradient indicates a higher correlation between a pair of features; hence better quality. The mean of all correlation score pairs is located in the top right corner.}
  \label{fig:correlations}
\end{figure*}

The fidelity level from the row perspective is evaluated by comparing feature correlations of the synthetic data produced by a variety of models. Figure \ref{fig:correlations} demonstrates an example of such correlations for the Credit Default dataset, spanning all features. A more intense color gradient signifies a stronger correlation between a pair of attributes, implying an enhanced quality of the synthesizer.

\section{Ablation Study}

We have observed the strong influence of normalization techniques when training the FinDiff model applied to numeric attributes of the Fund Holdings. Since this dataset contains mostly extremely skewed numeric attributes, the normalization technique has to be selected carefully. The \autoref{tab:scalers} presents a fidelity assessment of three different transformation or scaling techniques applied on Fund Holdings: Standard Scaler (zero mean and unit variance), Power Transformer ('yeo-johnson' method \citep{yeo_new_2000}) and
Quantile Transformer \footnote{\url{https://scikit-learn.org/stable/modules/generated/sklearn.preprocessing.QuantileTransformer.html}}. Notably, the Quantile Transformer emerges superior, achieving the highest fidelity scores in both metrics. Conversely, the Standard Scaler yields the lowest fidelity, suggesting a less accurate data replication. These results underline the influence of data scaling methods on the effectiveness of synthetic data generation.

\begin{table}[htbp]
    \centering
    \begin{tabular}{@{}lcc} 
    \toprule
         & \textbf{Fidelity Columns} & \textbf{Fidelity Row} \\
        \midrule
        Standard Scaler & 0.534 $\pm$ 0.01 & 0.824 $\pm$ 0.01 \\
        Power Transformer & 0.552 $\pm$ 0.03 & 0.889 $\pm$ 0.04 \\
        Quantile Transformer & \textbf{0.764} $\pm$ \textbf{0.01} & \textbf{0.949} $\pm$ \textbf{0.01} \\
        \bottomrule
    \end{tabular}
    \caption{Fidelity assessment of various transformation/scaling methods. The values represent the mean and standard deviation rom 5 experiments of random seeds.}
    \label{tab:scalers}
\end{table}

\section{Conclusion and Future work}
\label{sec:conclusion}

The present study introduces FinDif, a financial diffusion model, designed for generating synthetic financial tabular data aimed at enhancing downstream tasks. The model utilizes embedding encoding to address the challenges inherent to mixed-modality financial data. 
Furthermore, the model is equipped to generate conditional sampling, proving particularly beneficial for datasets with multiple classes. Evaluations of the model, undertaken from various perspectives including fidelity, privacy, and utility, suggest that the model outperforms all baseline models on public datasets and achieves notable results on proprietary datasets as per the set metrics. The potential for this model to serve as an effective tool within the financial regulatory environment is evident.

\begin{acks}
Due to the double-blind review, the acknowledgments are hidden.
\end{acks}

\bibliographystyle{ACM-Reference-Format}
\bibliography{bibliography}


\begin{thebibliography}{39}


\ifx \showCODEN    \undefined \def \showCODEN     #1{\unskip}     \fi
\ifx \showDOI      \undefined \def \showDOI       #1{#1}\fi
\ifx \showISBNx    \undefined \def \showISBNx     #1{\unskip}     \fi
\ifx \showISBNxiii \undefined \def \showISBNxiii  #1{\unskip}     \fi
\ifx \showISSN     \undefined \def \showISSN      #1{\unskip}     \fi
\ifx \showLCCN     \undefined \def \showLCCN      #1{\unskip}     \fi
\ifx \shownote     \undefined \def \shownote      #1{#1}          \fi
\ifx \showarticletitle \undefined \def \showarticletitle #1{#1}   \fi
\ifx \showURL      \undefined \def \showURL       {\relax}        \fi
\providecommand\bibfield[2]{#2}
\providecommand\bibinfo[2]{#2}
\providecommand\natexlab[1]{#1}
\providecommand\showeprint[2][]{arXiv:#2}

\bibitem[Hol({[n.\,d.]})]%
        {Holdings_hidden}
 \bibinfo{year}{[n.\,d.]}\natexlab{}.
\newblock \bibinfo{title}{Author names and paper title are redacted during the
  double-blind review}.
\newblock
\newblock


\bibitem[Assefa et~al\mbox{.}(2021)]%
        {synth_data_finance}
\bibfield{author}{\bibinfo{person}{Samuel~A. Assefa}, \bibinfo{person}{Danial
  Dervovic}, \bibinfo{person}{Mahmoud Mahfouz}, \bibinfo{person}{Robert~E.
  Tillman}, \bibinfo{person}{Prashant Reddy}, {and} \bibinfo{person}{Manuela
  Veloso}.} \bibinfo{year}{2021}\natexlab{}.
\newblock \showarticletitle{Generating Synthetic Data in Finance:
  Opportunities, Challenges and Pitfalls}. In
  \bibinfo{booktitle}{\emph{Proceedings of the First ACM International
  Conference on AI in Finance}} (New York, New York)
  \emph{(\bibinfo{series}{ICAIF '20})}. \bibinfo{publisher}{Association for
  Computing Machinery}, \bibinfo{address}{New York, NY, USA}, Article
  \bibinfo{articleno}{44}, \bibinfo{numpages}{8}~pages.
\newblock
\showISBNx{9781450375849}
\urldef\tempurl%
\url{https://doi.org/10.1145/3383455.3422554}
\showDOI{\tempurl}


\bibitem[Benesty et~al\mbox{.}(2009)]%
        {benesty2009pearson}
\bibfield{author}{\bibinfo{person}{Jacob Benesty}, \bibinfo{person}{Jingdong
  Chen}, \bibinfo{person}{Yiteng Huang}, {and} \bibinfo{person}{Israel Cohen}.}
  \bibinfo{year}{2009}\natexlab{}.
\newblock \showarticletitle{Pearson correlation coefficient}.
\newblock In \bibinfo{booktitle}{\emph{Noise reduction in speech processing}}.
  \bibinfo{publisher}{Springer}, \bibinfo{pages}{37--40}.
\newblock


\bibitem[Chen et~al\mbox{.}(2020)]%
        {chen2020wavegrad}
\bibfield{author}{\bibinfo{person}{Nanxin Chen}, \bibinfo{person}{Yu Zhang},
  \bibinfo{person}{Heiga Zen}, \bibinfo{person}{Ron~J. Weiss},
  \bibinfo{person}{Mohammad Norouzi}, {and} \bibinfo{person}{William Chan}.}
  \bibinfo{year}{2020}\natexlab{}.
\newblock \bibinfo{title}{WaveGrad: Estimating Gradients for Waveform
  Generation}.
\newblock
\newblock
\showeprint[arxiv]{2009.00713}~[eess.AS]


\bibitem[Croitoru et~al\mbox{.}(2023)]%
        {diffusion_models_survey}
\bibfield{author}{\bibinfo{person}{Florinel-Alin Croitoru},
  \bibinfo{person}{Vlad Hondru}, \bibinfo{person}{Radu~Tudor Ionescu}, {and}
  \bibinfo{person}{Mubarak Shah}.} \bibinfo{year}{2023}\natexlab{}.
\newblock \showarticletitle{Diffusion Models in Vision: A Survey}.
\newblock \bibinfo{journal}{\emph{IEEE Transactions on Pattern Analysis and
  Machine Intelligence}} (\bibinfo{year}{2023}), \bibinfo{pages}{1--20}.
\newblock
\urldef\tempurl%
\url{https://doi.org/10.1109/TPAMI.2023.3261988}
\showDOI{\tempurl}


\bibitem[Dogariu and Rebedea(2021)]%
        {dogariu2021synthetic}
\bibfield{author}{\bibinfo{person}{Mihai Dogariu} {and} \bibinfo{person}{Traian
  Rebedea}.} \bibinfo{year}{2021}\natexlab{}.
\newblock \showarticletitle{Synthetic Financial Time Series Generation using
  Generative Adversarial Networks}.
\newblock \bibinfo{journal}{\emph{ACM Transactions on Multimedia Computing,
  Communications, and Applications (TOMM)}} (\bibinfo{year}{2021}).
\newblock
\urldef\tempurl%
\url{https://doi.org/10.1145/3490354.3494393}
\showDOI{\tempurl}


\bibitem[Engelmann and Lessmann(2021)]%
        {engelmann2021conditional}
\bibfield{author}{\bibinfo{person}{Justin Engelmann} {and}
  \bibinfo{person}{Stefan Lessmann}.} \bibinfo{year}{2021}\natexlab{}.
\newblock \showarticletitle{Conditional Wasserstein GAN-based oversampling of
  tabular data for imbalanced learning}.
\newblock \bibinfo{journal}{\emph{Expert Systems with Applications}}
  \bibinfo{volume}{174} (\bibinfo{year}{2021}), \bibinfo{pages}{114582}.
\newblock


\bibitem[Fan et~al\mbox{.}(2020)]%
        {fan2020relational}
\bibfield{author}{\bibinfo{person}{Ju Fan}, \bibinfo{person}{Tongyu Liu},
  \bibinfo{person}{Guoliang Li}, \bibinfo{person}{Junyou Chen},
  \bibinfo{person}{Yuwei Shen}, {and} \bibinfo{person}{Xiaoyong Du}.}
  \bibinfo{year}{2020}\natexlab{}.
\newblock \showarticletitle{Relational data synthesis using generative
  adversarial networks: A design space exploration}.
\newblock \bibinfo{journal}{\emph{arXiv preprint arXiv:2008.12763}}
  (\bibinfo{year}{2020}).
\newblock


\bibitem[Fonseca and Bacao(2023)]%
        {fonseca_tabular_2023}
\bibfield{author}{\bibinfo{person}{Joao Fonseca} {and}
  \bibinfo{person}{Fernando Bacao}.} \bibinfo{year}{2023}\natexlab{}.
\newblock \showarticletitle{Tabular and latent space synthetic data generation:
  a literature review}.
\newblock \bibinfo{journal}{\emph{Journal of Big Data}} \bibinfo{volume}{10},
  \bibinfo{number}{1} (\bibinfo{date}{July} \bibinfo{year}{2023}),
  \bibinfo{pages}{115}.
\newblock
\showISSN{2196-1115}
\urldef\tempurl%
\url{https://doi.org/10.1186/s40537-023-00792-7}
\showDOI{\tempurl}


\bibitem[Gao et~al\mbox{.}(2023)]%
        {gao2023difformer}
\bibfield{author}{\bibinfo{person}{Zhujin Gao}, \bibinfo{person}{Junliang Guo},
  \bibinfo{person}{Xu Tan}, \bibinfo{person}{Yongxin Zhu},
  \bibinfo{person}{Fang Zhang}, \bibinfo{person}{Jiang Bian}, {and}
  \bibinfo{person}{Linli Xu}.} \bibinfo{year}{2023}\natexlab{}.
\newblock \bibinfo{title}{Difformer: Empowering Diffusion Models on the
  Embedding Space for Text Generation}.
\newblock
\newblock
\showeprint[arxiv]{2212.09412}~[cs.CL]


\bibitem[Glorot and Bengio(2010)]%
        {Glorot10understandingthe}
\bibfield{author}{\bibinfo{person}{Xavier Glorot} {and} \bibinfo{person}{Yoshua
  Bengio}.} \bibinfo{year}{2010}\natexlab{}.
\newblock \showarticletitle{Understanding the difficulty of training deep
  feedforward neural networks}. In \bibinfo{booktitle}{\emph{Proceedings of the
  Thirteenth International Conference on Artificial Intelligence and
  Statistics}} \emph{(\bibinfo{series}{Proceedings of Machine Learning
  Research}, Vol.~\bibinfo{volume}{9})},
  \bibfield{editor}{\bibinfo{person}{Yee~Whye Teh} {and} \bibinfo{person}{Mike
  Titterington}} (Eds.). \bibinfo{publisher}{PMLR}, \bibinfo{address}{Chia
  Laguna Resort, Sardinia, Italy}, \bibinfo{pages}{249--256}.
\newblock
\urldef\tempurl%
\url{https://proceedings.mlr.press/v9/glorot10a.html}
\showURL{%
\tempurl}


\bibitem[Ho et~al\mbox{.}(2020)]%
        {ho2020denoising}
\bibfield{author}{\bibinfo{person}{Jonathan Ho}, \bibinfo{person}{Ajay Jain},
  {and} \bibinfo{person}{Pieter Abbeel}.} \bibinfo{year}{2020}\natexlab{}.
\newblock \bibinfo{title}{Denoising Diffusion Probabilistic Models}.
\newblock
\newblock
\showeprint[arxiv]{2006.11239}~[cs.LG]


\bibitem[Hoogeboom et~al\mbox{.}(2021)]%
        {argmax_flows}
\bibfield{author}{\bibinfo{person}{Emiel Hoogeboom}, \bibinfo{person}{Didrik
  Nielsen}, \bibinfo{person}{Priyank Jaini}, \bibinfo{person}{Patrick
  Forr\'{e}}, {and} \bibinfo{person}{Max Welling}.}
  \bibinfo{year}{2021}\natexlab{}.
\newblock \showarticletitle{Argmax Flows and Multinomial Diffusion: Learning
  Categorical Distributions}. In \bibinfo{booktitle}{\emph{Advances in Neural
  Information Processing Systems}},
  \bibfield{editor}{\bibinfo{person}{M.~Ranzato},
  \bibinfo{person}{A.~Beygelzimer}, \bibinfo{person}{Y.~Dauphin},
  \bibinfo{person}{P.S. Liang}, {and} \bibinfo{person}{J.~Wortman Vaughan}}
  (Eds.), Vol.~\bibinfo{volume}{34}. \bibinfo{publisher}{Curran Associates,
  Inc.}, \bibinfo{pages}{12454--12465}.
\newblock
\urldef\tempurl%
\url{https://proceedings.neurips.cc/paper_files/paper/2021/file/67d96d458abdef21792e6d8e590244e7-Paper.pdf}
\showURL{%
\tempurl}


\bibitem[Kingma and Ba(2015)]%
        {adam}
\bibfield{author}{\bibinfo{person}{Diederik~P. Kingma} {and}
  \bibinfo{person}{Jimmy Ba}.} \bibinfo{year}{2015}\natexlab{}.
\newblock \showarticletitle{Adam: {A} Method for Stochastic Optimization}. In
  \bibinfo{booktitle}{\emph{3rd International Conference on Learning
  Representations, {ICLR} 2015, San Diego, CA, USA, May 7-9, 2015, Conference
  Track Proceedings}}, \bibfield{editor}{\bibinfo{person}{Yoshua Bengio} {and}
  \bibinfo{person}{Yann LeCun}} (Eds.).
\newblock
\urldef\tempurl%
\url{http://arxiv.org/abs/1412.6980}
\showURL{%
\tempurl}


\bibitem[Kingma and Welling(2014)]%
        {vae2014}
\bibfield{author}{\bibinfo{person}{Diederik~P. Kingma} {and}
  \bibinfo{person}{Max Welling}.} \bibinfo{year}{2014}\natexlab{}.
\newblock \showarticletitle{Auto-Encoding Variational Bayes}. In
  \bibinfo{booktitle}{\emph{2nd International Conference on Learning
  Representations, {ICLR} 2014, Banff, AB, Canada, April 14-16, 2014,
  Conference Track Proceedings}}, \bibfield{editor}{\bibinfo{person}{Yoshua
  Bengio} {and} \bibinfo{person}{Yann LeCun}} (Eds.).
\newblock
\urldef\tempurl%
\url{http://arxiv.org/abs/1312.6114}
\showURL{%
\tempurl}


\bibitem[Kong et~al\mbox{.}(2021)]%
        {kong2021diffwave}
\bibfield{author}{\bibinfo{person}{Zhifeng Kong}, \bibinfo{person}{Wei Ping},
  \bibinfo{person}{Jiaji Huang}, \bibinfo{person}{Kexin Zhao}, {and}
  \bibinfo{person}{Bryan Catanzaro}.} \bibinfo{year}{2021}\natexlab{}.
\newblock \showarticletitle{DiffWave: A Versatile Diffusion Model for Audio
  Synthesis}. In \bibinfo{booktitle}{\emph{ICLR}}.
\newblock
\urldef\tempurl%
\url{https://openreview.net/forum?id=a-xFK8Ymz5J}
\showURL{%
\tempurl}


\bibitem[Kotelnikov et~al\mbox{.}(2022)]%
        {kotelnikov2022tabddpm}
\bibfield{author}{\bibinfo{person}{Akim Kotelnikov}, \bibinfo{person}{Dmitry
  Baranchuk}, \bibinfo{person}{Ivan Rubachev}, {and} \bibinfo{person}{Artem
  Babenko}.} \bibinfo{year}{2022}\natexlab{}.
\newblock \bibinfo{title}{TabDDPM: Modelling Tabular Data with Diffusion
  Models}.
\newblock
\newblock
\showeprint[arxiv]{2209.15421}~[cs.LG]


\bibitem[Li et~al\mbox{.}(2022)]%
        {li2022diffusionlm}
\bibfield{author}{\bibinfo{person}{Xiang~Lisa Li}, \bibinfo{person}{John
  Thickstun}, \bibinfo{person}{Ishaan Gulrajani}, \bibinfo{person}{Percy
  Liang}, {and} \bibinfo{person}{Tatsunori Hashimoto}.}
  \bibinfo{year}{2022}\natexlab{}.
\newblock \showarticletitle{Diffusion-{LM} Improves Controllable Text
  Generation}. In \bibinfo{booktitle}{\emph{Advances in Neural Information
  Processing Systems}}.
\newblock


\bibitem[Massey(1951)]%
        {kolmogorov_1951}
\bibfield{author}{\bibinfo{person}{F.~J. Massey}.}
  \bibinfo{year}{1951}\natexlab{}.
\newblock \showarticletitle{The {K}olmogorov-{S}mirnov test for goodness of
  fit}.
\newblock \bibinfo{journal}{\emph{J. Amer. Statist. Assoc.}}
  \bibinfo{volume}{46}, \bibinfo{number}{253} (\bibinfo{year}{1951}),
  \bibinfo{pages}{68--78}.
\newblock


\bibitem[Ni et~al\mbox{.}(2021)]%
        {ni2021sigwasserstein}
\bibfield{author}{\bibinfo{person}{Hao Ni}, \bibinfo{person}{Lukasz Szpruch},
  \bibinfo{person}{Marc Sabate-Vidales}, \bibinfo{person}{Baoren Xiao},
  \bibinfo{person}{Magnus Wiese}, {and} \bibinfo{person}{Shujian Liao}.}
  \bibinfo{year}{2021}\natexlab{}.
\newblock \bibinfo{title}{Sig-Wasserstein GANs for Time Series Generation}.
\newblock
\newblock
\showeprint[arxiv]{2111.01207}~[cs.LG]


\bibitem[Ni et~al\mbox{.}(2020)]%
        {ni2020conditional}
\bibfield{author}{\bibinfo{person}{Hao Ni}, \bibinfo{person}{Lukasz Szpruch},
  \bibinfo{person}{Magnus Wiese}, \bibinfo{person}{Shujian Liao}, {and}
  \bibinfo{person}{Baoren Xiao}.} \bibinfo{year}{2020}\natexlab{}.
\newblock \bibinfo{title}{Conditional Sig-Wasserstein GANs for Time Series
  Generation}.
\newblock
\newblock
\showeprint[arxiv]{2006.05421}~[cs.LG]


\bibitem[Nichol and Dhariwal(2021)]%
        {nichol2021improved}
\bibfield{author}{\bibinfo{person}{Alex Nichol} {and} \bibinfo{person}{Prafulla
  Dhariwal}.} \bibinfo{year}{2021}\natexlab{}.
\newblock \bibinfo{title}{Improved Denoising Diffusion Probabilistic Models}.
\newblock
\newblock
\showeprint[arxiv]{2102.09672}~[cs.LG]


\bibitem[Ouyang et~al\mbox{.}(2023)]%
        {ouyang2023missdiff}
\bibfield{author}{\bibinfo{person}{Yidong Ouyang}, \bibinfo{person}{Liyan Xie},
  \bibinfo{person}{Chongxuan Li}, {and} \bibinfo{person}{Guang Cheng}.}
  \bibinfo{year}{2023}\natexlab{}.
\newblock \showarticletitle{MissDiff: Training Diffusion Models on Tabular Data
  with Missing Values}. In \bibinfo{booktitle}{\emph{ICML 2023 Workshop on
  Structured Probabilistic Inference {\&} Generative Modeling}}.
\newblock
\urldef\tempurl%
\url{https://openreview.net/forum?id=S435pkeAdT}
\showURL{%
\tempurl}


\bibitem[Paszke and Gross(2019)]%
        {pytorch}
\bibfield{author}{\bibinfo{person}{Adam Paszke} {and} \bibinfo{person}{Sam
  et~al. Gross}.} \bibinfo{year}{2019}\natexlab{}.
\newblock \showarticletitle{PyTorch: An Imperative Style, High-Performance Deep
  Learning Library}.
\newblock In \bibinfo{booktitle}{\emph{Advances in Neural Information
  Processing Systems 32}}, \bibfield{editor}{\bibinfo{person}{H.~Wallach},
  \bibinfo{person}{H.~Larochelle}, \bibinfo{person}{A.~Beygelzimer},
  \bibinfo{person}{F.~d'~Alch\'{e}-Buc}, \bibinfo{person}{E.~Fox}, {and}
  \bibinfo{person}{R.~Garnett}} (Eds.). \bibinfo{publisher}{Curran Associates,
  Inc.}, \bibinfo{pages}{8024--8035}.
\newblock
\urldef\tempurl%
\url{http://papers.neurips.cc/paper/9015-pytorch-an-imperative-style-high-performance-deep-learning-library.pdf}
\showURL{%
\tempurl}


\bibitem[Patki et~al\mbox{.}(2016)]%
        {SDV}
\bibfield{author}{\bibinfo{person}{Neha Patki}, \bibinfo{person}{Roy Wedge},
  {and} \bibinfo{person}{Kalyan Veeramachaneni}.}
  \bibinfo{year}{2016}\natexlab{}.
\newblock \showarticletitle{The Synthetic data vault}. In
  \bibinfo{booktitle}{\emph{IEEE International Conference on Data Science and
  Advanced Analytics (DSAA)}}. \bibinfo{pages}{399--410}.
\newblock
\urldef\tempurl%
\url{https://doi.org/10.1109/DSAA.2016.49}
\showDOI{\tempurl}


\bibitem[Rombach et~al\mbox{.}(2022)]%
        {rombach2022highresolution}
\bibfield{author}{\bibinfo{person}{Robin Rombach}, \bibinfo{person}{Andreas
  Blattmann}, \bibinfo{person}{Dominik Lorenz}, \bibinfo{person}{Patrick
  Esser}, {and} \bibinfo{person}{Björn Ommer}.}
  \bibinfo{year}{2022}\natexlab{}.
\newblock \bibinfo{title}{High-Resolution Image Synthesis with Latent Diffusion
  Models}.
\newblock
\newblock
\showeprint[arxiv]{2112.10752}~[cs.CV]


\bibitem[Schreyer et~al\mbox{.}(2019)]%
        {schreyer2019adversarial}
\bibfield{author}{\bibinfo{person}{Marco Schreyer}, \bibinfo{person}{Timur
  Sattarov}, \bibinfo{person}{Bernd Reimer}, {and} \bibinfo{person}{Damian
  Borth}.} \bibinfo{year}{2019}\natexlab{}.
\newblock \bibinfo{title}{Adversarial Learning of Deepfakes in Accounting}.
\newblock
\newblock
\showeprint[arxiv]{1910.03810}~[cs.LG]


\bibitem[Sohl-Dickstein et~al\mbox{.}(2015)]%
        {sohldickstein2015deep}
\bibfield{author}{\bibinfo{person}{Jascha Sohl-Dickstein},
  \bibinfo{person}{Eric~A. Weiss}, \bibinfo{person}{Niru Maheswaranathan},
  {and} \bibinfo{person}{Surya Ganguli}.} \bibinfo{year}{2015}\natexlab{}.
\newblock \bibinfo{title}{Deep Unsupervised Learning using Nonequilibrium
  Thermodynamics}.
\newblock
\newblock
\showeprint[arxiv]{1503.03585}~[cs.LG]


\bibitem[Song et~al\mbox{.}(2022)]%
        {song2022denoising}
\bibfield{author}{\bibinfo{person}{Jiaming Song}, \bibinfo{person}{Chenlin
  Meng}, {and} \bibinfo{person}{Stefano Ermon}.}
  \bibinfo{year}{2022}\natexlab{}.
\newblock \bibinfo{title}{Denoising Diffusion Implicit Models}.
\newblock
\newblock
\showeprint[arxiv]{2010.02502}~[cs.LG]


\bibitem[Strudel et~al\mbox{.}(2022)]%
        {strudel2022selfconditioned}
\bibfield{author}{\bibinfo{person}{Robin Strudel}, \bibinfo{person}{Corentin
  Tallec}, \bibinfo{person}{Florent Altché}, \bibinfo{person}{Yilun Du},
  \bibinfo{person}{Yaroslav Ganin}, \bibinfo{person}{Arthur Mensch},
  \bibinfo{person}{Will Grathwohl}, \bibinfo{person}{Nikolay Savinov},
  \bibinfo{person}{Sander Dieleman}, \bibinfo{person}{Laurent Sifre}, {and}
  \bibinfo{person}{Rémi Leblond}.} \bibinfo{year}{2022}\natexlab{}.
\newblock \bibinfo{title}{Self-conditioned Embedding Diffusion for Text
  Generation}.
\newblock
\newblock
\showeprint[arxiv]{2211.04236}~[cs.CL]


\bibitem[Vardhan and Kok(2020)]%
        {vardhan2020generating}
\bibfield{author}{\bibinfo{person}{L~Vivek~Harsha Vardhan} {and}
  \bibinfo{person}{Stanley Kok}.} \bibinfo{year}{2020}\natexlab{}.
\newblock \showarticletitle{Generating privacy-preserving synthetic tabular
  data using oblivious variational autoencoders}. In
  \bibinfo{booktitle}{\emph{Proceedings of the Workshop on Economics of Privacy
  and Data Labor at the 37 th International Conference on Machine Learning}}.
\newblock


\bibitem[Wan et~al\mbox{.}(2017)]%
        {vae_tab1}
\bibfield{author}{\bibinfo{person}{Zhiqiang Wan}, \bibinfo{person}{Yazhou
  Zhang}, {and} \bibinfo{person}{Haibo He}.} \bibinfo{year}{2017}\natexlab{}.
\newblock \showarticletitle{Variational autoencoder based synthetic data
  generation for imbalanced learning}. In \bibinfo{booktitle}{\emph{2017 IEEE
  Symposium Series on Computational Intelligence (SSCI)}}.
  \bibinfo{pages}{1--7}.
\newblock


\bibitem[Wiese et~al\mbox{.}(2019)]%
        {wiese2019quant}
\bibfield{author}{\bibinfo{person}{Jannes Wiese}, \bibinfo{person}{Andre
  Knobloch}, \bibinfo{person}{Walther Kretschmer}, {and}
  \bibinfo{person}{Thomas Huschto}.} \bibinfo{year}{2019}\natexlab{}.
\newblock \showarticletitle{Quant GANs: Deep Generation of Financial Time
  Series}.
\newblock \bibinfo{journal}{\emph{arXiv preprint arXiv:1907.06673}}
  (\bibinfo{year}{2019}).
\newblock


\bibitem[Wu et~al\mbox{.}(2023)]%
        {vae_fin1}
\bibfield{author}{\bibinfo{person}{Jinhong Wu}, \bibinfo{person}{Konstantinos
  Plataniotis}, \bibinfo{person}{Lucy Liu}, \bibinfo{person}{Ehsan Amjadian},
  {and} \bibinfo{person}{Yuri Lawryshyn}.} \bibinfo{year}{2023}\natexlab{}.
\newblock \showarticletitle{Interpretation for Variational Autoencoder Used to
  Generate Financial Synthetic Tabular Data}.
\newblock \bibinfo{journal}{\emph{Algorithms}} \bibinfo{volume}{16},
  \bibinfo{number}{2} (\bibinfo{year}{2023}).
\newblock
\showISSN{1999-4893}
\urldef\tempurl%
\url{https://www.mdpi.com/1999-4893/16/2/121}
\showURL{%
\tempurl}


\bibitem[Xu et~al\mbox{.}(2019)]%
        {tvae_ctgan}
\bibfield{author}{\bibinfo{person}{Lei Xu}, \bibinfo{person}{Maria
  Skoularidou}, \bibinfo{person}{Alfredo Cuesta-Infante}, {and}
  \bibinfo{person}{Kalyan Veeramachaneni}.} \bibinfo{year}{2019}\natexlab{}.
\newblock \showarticletitle{Modeling tabular data using conditional gan}.
\newblock \bibinfo{journal}{\emph{NeurIPS}}  \bibinfo{volume}{32}
  (\bibinfo{year}{2019}).
\newblock


\bibitem[Yeh and Lien(2009)]%
        {credit_default}
\bibfield{author}{\bibinfo{person}{I.~C. Yeh} {and} \bibinfo{person}{C.~H.
  Lien}.} \bibinfo{year}{2009}\natexlab{}.
\newblock \bibinfo{title}{The comparisons of data mining techniques for the
  predictive accuracy of probability of default of credit card clients.}
\newblock
\newblock


\bibitem[Yeo and Johnson(2000)]%
        {yeo_new_2000}
\bibfield{author}{\bibinfo{person}{In‐Kwon Yeo} {and}
  \bibinfo{person}{Richard~A. Johnson}.} \bibinfo{year}{2000}\natexlab{}.
\newblock \showarticletitle{A new family of power transformations to improve
  normality or symmetry}.
\newblock \bibinfo{journal}{\emph{Biometrika}} \bibinfo{volume}{87},
  \bibinfo{number}{4} (\bibinfo{date}{Dec.} \bibinfo{year}{2000}),
  \bibinfo{pages}{954--959}.
\newblock
\showISSN{0006-3444}
\urldef\tempurl%
\url{https://doi.org/10.1093/biomet/87.4.954}
\showDOI{\tempurl}
\newblock
\shownote{\_eprint:
  https://academic.oup.com/biomet/article-pdf/87/4/954/633221/870954.pdf}.


\bibitem[Zhang et~al\mbox{.}(2023)]%
        {zhang2023texttoimage}
\bibfield{author}{\bibinfo{person}{Chenshuang Zhang}, \bibinfo{person}{Chaoning
  Zhang}, \bibinfo{person}{Mengchun Zhang}, {and} \bibinfo{person}{In~So
  Kweon}.} \bibinfo{year}{2023}\natexlab{}.
\newblock \bibinfo{title}{Text-to-image Diffusion Models in Generative AI: A
  Survey}.
\newblock
\newblock
\showeprint[arxiv]{2303.07909}~[cs.CV]


\bibitem[Zou et~al\mbox{.}(2023)]%
        {zou2023survey}
\bibfield{author}{\bibinfo{person}{Hao Zou}, \bibinfo{person}{Zae~Myung Kim},
  {and} \bibinfo{person}{Dongyeop Kang}.} \bibinfo{year}{2023}\natexlab{}.
\newblock \bibinfo{title}{A Survey of Diffusion Models in Natural Language
  Processing}.
\newblock
\newblock
\showeprint[arxiv]{2305.14671}~[cs.CL]


\end{thebibliography}

\appendix

\end{document}